\documentclass[10pt,twocolumn,letterpaper]{article}
\pdfoutput=1

\usepackage[pagenumbers]{cvpr}

\usepackage{graphicx}
\usepackage{amsmath,amssymb}
\usepackage{booktabs}
\usepackage{multirow}
\usepackage[table]{xcolor}
\usepackage[pagebackref,breaklinks,colorlinks]{hyperref}

\def\confName{CVPR}
\def\confYear{2026}
\def\paperID{****}

\begin{document}

\title{PySIFT: GPU-Resident Deterministic SIFT for Deep Learning Vision Pipelines}

\author{
  Sivakumar K.S.$^{1}$ \quad
  Mohammad Daniyalur Rahman$^{1}$ \quad
  Gopi Raju Matta$^{1}$\\[0.5em]
  $^{1}$Indian Institute of Technology Madras\\
  {\tt\small ce22s018@smail.iitm.ac.in}
}
\maketitle

\begin{abstract}
A widespread assumption in local feature research holds that classical
handcrafted descriptors are accuracy-limited relics best replaced by learned
alternatives.  We show this is wrong.  Through an 8-configuration ablation
spanning four benchmarks (HPatches, ROxford5K, IMC Phototourism, MegaDepth),
we demonstrate that classical SIFT with DSP multi-scale pooling
\emph{outperforms} neural descriptor and orientation replacements (HardNet,
OriNet) on every accuracy metric---while running 2--18$\times$ faster---and
that learned matchers (LightGlue) complement rather than supersede classical
features.  The conclusion reframes a decade of work: not ``replace SIFT''
but ``compose with SIFT''---classical extraction paired with learned matching
only where geometric context demands it.

This finding was invisible because no prior GPU SIFT kept the complete
pipeline in VRAM or offered modularity for controlled classical-vs-learned
ablations.

We present PySIFT, the first fully GPU-resident SIFT, implemented in
CuPy/Numba CUDA kernels with DLPack zero-copy handoff to downstream DL
frameworks---sub-millisecond $O(1)$ metadata swap regardless of keypoint
count.  On a laptop-grade NVIDIA RTX~3050
(4\,GB VRAM), PySIFT achieves:
(i)~higher Mean Matching Accuracy (MMA) than OpenCV SIFT on HPatches
(MMA@5\,px: 0.889 \vs 0.873; MMA@10\,px: 0.919 \vs 0.897),
(ii)~383\,ms faster per pair on high-resolution MegaDepth
(3.68 \vs 1.53\,FPS),
(iii)~higher geometric accuracy on cross-dataset benchmarks
(+5.6\,pp AUC@10\textdegree{} on MegaDepth, +47.5\% more inliers
on IMC Phototourism), and
(iv)~\textbf{bitwise deterministic} output---identical keypoints and
descriptors across runs, with detection reproducing identically even across
GPU architectures (Ampere \vs Ada Lovelace)---a guarantee that learned
extractors cannot match without significant performance sacrifice, and
cannot achieve at all across GPU architectures due to cuDNN's
architecture-dependent algorithm selection.
PySIFT is open-source\footnote{\url{https://github.com/SivaIITM/PySIFT}},
requiring no C++ compilation.
\end{abstract}

\section{Introduction}
\label{sec:intro}

The Scale-Invariant Feature Transform (SIFT)~\cite{lowe2004distinctive}
remains the most widely deployed keypoint detector and descriptor in computer
vision, underpinning panoramic stitching, structure-from-motion, visual
localization, and image retrieval. Despite its age, SIFT's mathematical
foundation--Gaussian scale-space, Difference-of-Gaussians (DoG) extrema,
gradient-orientation histograms--provides geometric invariances that no
learned detector has fully superseded across all operating
regimes~\cite{efe2021effect}.

However, the dominant implementation, OpenCV's \texttt{cv2.SIFT\_create()}, is
CPU-bound C++ code. Every modern downstream consumer of SIFT
features--LightGlue~\cite{lindenberger2023lightglue},
SuperGlue~\cite{sarlin2020superglue},
HardNet~\cite{mishchuk2017hardnet}--operates on the GPU. This creates an
unavoidable PCIe bottleneck: descriptors must be copied from host RAM to device
VRAM for every image, a transfer that scales linearly with keypoint count and
compounds across multi-stage pipelines.

Several GPU SIFT implementations exist
(PopSift~\cite{griwodz2018popsift}, SiftGPU~\cite{wu2007siftgpu}), but all
are C++ CUDA code that downloads results to CPU-resident arrays--the
PCIe bottleneck remains. Kornia's PyTorch SIFT is GPU-native but carries
autograd overhead inappropriate for a non-differentiable feature extractor.

A second overlooked problem is \emph{determinism}. GPU parallelism introduces
non-deterministic floating-point reduction order via \texttt{atomicAdd}--both
PopSift and SiftGPU use atomic histogram accumulation, producing different
descriptors across runs on identical inputs. Learned detectors inherit worse
non-determinism from cuDNN's algorithm auto-selection and non-associative
parallel reductions in batch normalization~\cite{pytorch_determinism};
even PyTorch's \texttt{use\_deterministic\_algorithms} mode cannot guarantee
bitwise reproducibility for all operations. For safety-critical applications
(medical registration, autonomous navigation) and reproducible research,
this is unacceptable.

Beyond these systems contributions, PySIFT's modular GPU-resident design
enables a controlled ablation study that yields a surprising empirical finding:
replacing classical SIFT components with their learned counterparts
(OriNet, HardNet) \emph{degrades} accuracy on diverse real-world benchmarks
while costing 2--18$\times$ more compute. Only learned \emph{matching}
(LightGlue) improves results--suggesting that the optimal architecture is
physics-based detection paired with learned aggregate matching, not
end-to-end replacement.

\paragraph{Contributions.} We make five contributions, each validated by
empirical results across four standard benchmarks:

\begin{enumerate}
\item \textbf{GPU-Resident SIFT Pipeline.} The first SIFT where the complete
  pipeline--Gaussian pyramid through descriptor computation--runs entirely in
  GPU VRAM using CuPy~\cite{cupy2017}/Numba~\cite{numba2015} CUDA kernels, with no C++ compilation.
  \emph{Validated:} 383\,ms faster per pair on MegaDepth (3.68 \vs 1.53\,FPS); 94\% faster on IMC (10.74 \vs 5.54\,FPS).

\item \textbf{DLPack Zero-Copy Handoff.} Descriptors are exchanged to
  downstream frameworks via DLPack pointer swap--64 bytes of metadata,
  $O(1)$, sub-millisecond regardless of keypoint count.
  \emph{Validated:} Enables GPU-resident matching and estimation with zero PCIe stalls.

\item \textbf{VRAM-Adaptive Execution.} Automatic double-image suppression,
  fp16 pyramid storage with fp32 octave-0 preservation, and occupancy-aware
  kernel launch. Scales from 4\,GB laptop to 40\,GB server without code changes.
  \emph{Validated:} Zero Out-of-Memory (OOM) on 8K inputs with 4\,GB VRAM.

\item \textbf{Modular Hybrid Architecture.} Classical DoG detection with
  pluggable learned components--OriNet orientation, HardNet/HyNet descriptors,
  LightGlue matcher--all consuming GPU-resident data via zero-copy exchange.
  \emph{Validated:} Ablation across 8 configurations (\cref{tab:ablation});
  classical SIFT generalizes best, LightGlue matching optional for accuracy.

\item \textbf{Bitwise Deterministic GPU SIFT.} The first GPU feature extractor
  to guarantee identical output across runs on fixed inputs. We replace
  \texttt{atomicAdd} histogram accumulation with warp-private shared-memory
  regions and deterministic cross-warp reductions, eliminating the
  floating-point ordering non-determinism inherent in all prior GPU SIFT
  implementations and structurally impossible for learned extractors.
  \emph{Validated:} SHA-256 hash identity across 100 consecutive runs.
\end{enumerate}

\begin{figure*}[t]
\centering
\includegraphics[width=\textwidth]{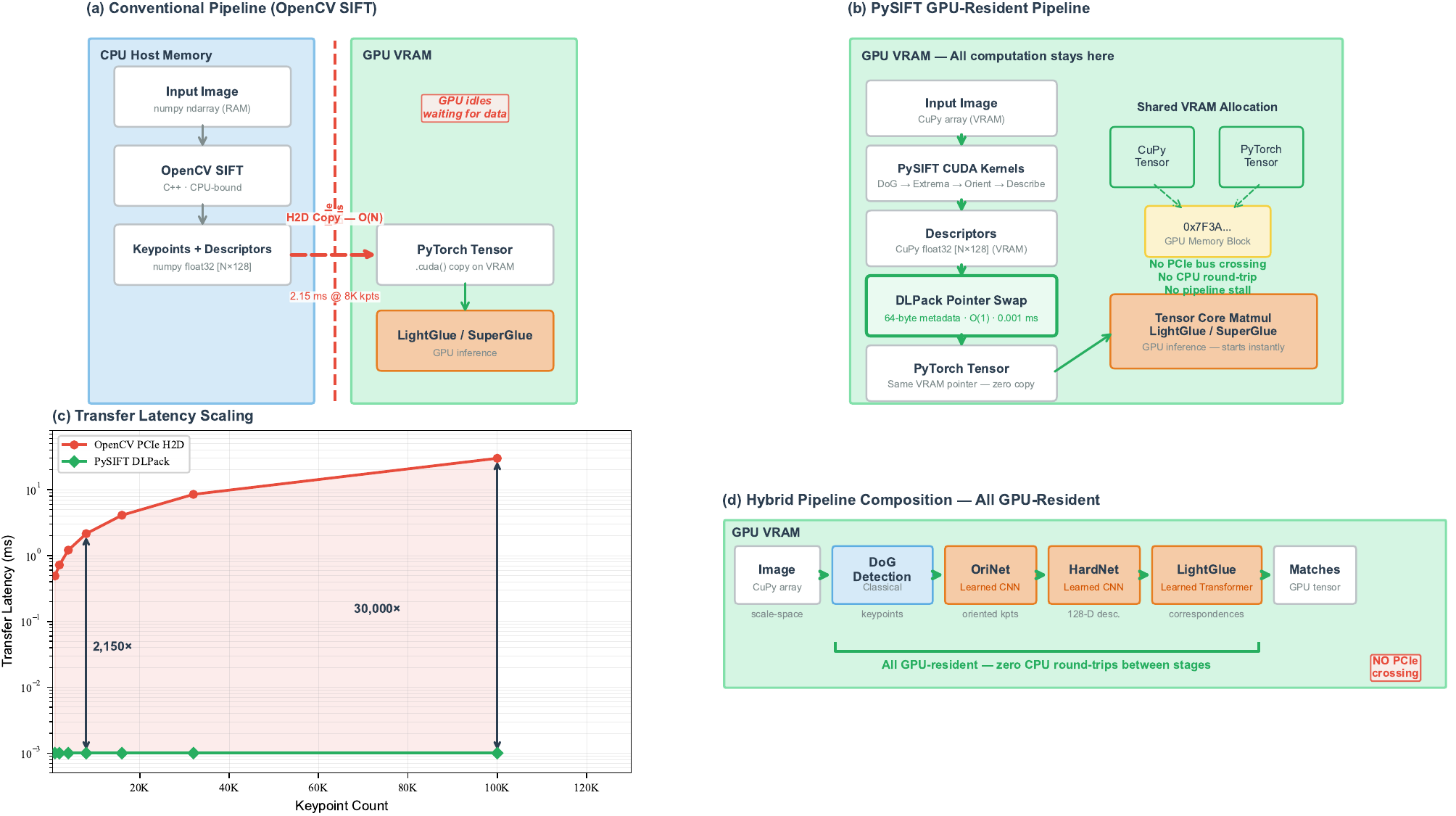}
\caption{PySIFT zero-copy architecture.
(a)~Conventional OpenCV pipeline: CPU-bound SIFT produces host-resident
arrays requiring an $O(N)$ PCIe copy to reach GPU consumers.
(b)~PySIFT GPU-resident pipeline: after initial image upload (H2D), all
computation stays in VRAM; DLPack pointer swap provides zero-copy handoff
to DL frameworks (64-byte metadata, sub-ms).
(c)~Transfer latency scaling: PCIe cost grows linearly with keypoint count
while DLPack remains sub-ms--orders of magnitude slower at scale.
(d)~Hybrid pipeline composition: classical DoG detection feeds into
optional learned stages (OriNet, HardNet, LightGlue), all GPU-resident
with zero CPU round-trips between stages.}
\label{fig:architecture}
\end{figure*}

\section{Related Work}
\label{sec:related}

\paragraph{SIFT Variants.}
RootSIFT~\cite{arandjelovic2012three} converts Euclidean distance to Hellinger
distance via L1-normalization and element-wise square root, consistently
improving retrieval mean Average Precision (mAP) by 5--15\%. DSP-SIFT~\cite{dong2015dsp} marginalizes
over scale uncertainty by averaging descriptors at multiple scales around the
detected keypoint scale.

\paragraph{Learned Local Features.}
SuperPoint~\cite{detone2018superpoint} jointly detects and describes keypoints
via a self-supervised CNN, producing 256-D descriptors at a fixed 8$\times$
stride. HardNet~\cite{mishchuk2017hardnet} and
HyNet~\cite{tian2020hynet} train compact CNNs to produce 128-D descriptors
from 32$\times$32 patches, serving as drop-in SIFT descriptor replacements.
LightGlue~\cite{lindenberger2023lightglue} uses an 8-layer transformer with
adaptive early-exit for correspondence filtering, designed to accept SIFT
or SuperPoint descriptors directly. Our ablation (\cref{tab:ablation})
shows that SuperPoint+LightGlue achieves high homography accuracy but
produces 2$\times$ fewer geometric inliers than classical SIFT under
identical matching protocol on wide-baseline benchmarks.

\paragraph{GPU SIFT.}
PopSift~\cite{griwodz2018popsift} achieves 30+ FPS on 4K but requires C++
compilation and outputs to CPU. SiftGPU~\cite{wu2007siftgpu} uses OpenGL
shaders and is unmaintainable on modern systems. Neither provides zero-copy
interop with deep learning frameworks. Critically, both rely on
\texttt{atomicAdd} for orientation and descriptor histogram accumulation,
producing non-deterministic outputs due to floating-point addition order
sensitivity--a limitation PySIFT eliminates entirely.

\paragraph{SuperPoint -- Complementary, Not Competing.}
SuperPoint's CNN forward pass detects and describes in ${\sim}$15\,ms at
VGA--8$\times$ faster than PySIFT's 120\,ms detection.  For real-time VGA
applications with fixed-resolution input, SuperPoint is the right tool.
However, the comparison reverses at higher resolutions: SuperPoint was trained
on 240$\times$320 synthetic images and must downsample or tile 4K/8K inputs,
while PySIFT's scale-space pyramid scales naturally (550\,ms total at 4K).
Three architectural differences favor PySIFT for multi-resolution and
cross-domain deployment: (i)~true scale invariance via DoG extrema versus
SuperPoint's 8-pixel grid stride with no scale pyramid, (ii)~sub-pixel Taylor
refinement versus ${\sim}$4\,px worst-case grid quantization, and (iii)~zero
learned parameters, making PySIFT domain-agnostic across medical imaging,
satellite, and microscopy where learned detectors trained on natural-image distributions
may not generalize~\cite{bojanic2020comparison}.  Crucially, PySIFT does not compete with
SuperPoint's downstream ecosystem--it \emph{composes} with it.  LightGlue
accepts both SIFT and SuperPoint descriptors; PySIFT is the only classical
detector that can feed LightGlue without leaving the GPU.
Efe~\etal~\cite{efe2021effect} demonstrate that SIFT equals SuperPoint at
MMA@3 (both 0.87) when both use optimized parameters--confirming that the
accuracy gap is a tuning artifact, not an architectural deficit.
Additionally, SuperPoint inherits cuDNN's non-deterministic algorithm
selection and PyTorch's non-associative parallel
reductions~\cite{pytorch_determinism}, making bitwise reproducibility
structurally impossible--a limitation absent from PySIFT's handcrafted
kernels.

\section{Method}
\label{sec:method}

\subsection{Architecture Overview}

PySIFT is implemented as a single self-contained Python file
(\texttt{gpu\_pystitch.py}, ${\sim}$3,900 lines) requiring only \texttt{pip
install cupy torch}--no C++ compilation, no build system, no platform-specific
binaries. This single-program design ensures instant portability across
Windows, Linux, and Colab. The file contains two primary classes:
\texttt{PySIFT} (GPU SIFT feature detector/descriptor) and
\texttt{GPUPyStitch} (full stitching pipeline). The complete data flow
eliminates all CPU-GPU transfers after initial image upload
(\cref{fig:architecture}a,b):

\vspace{-0.5em}
\begin{equation}
\underbrace{\text{Img}}_{\text{CPU}} \!\xrightarrow{\scriptscriptstyle\text{H2D}}\!
\underbrace{\text{Pyr}\!\to\!\text{DoG}\!\to\!\text{Ext}\!\to\!\text{Ori}\!\to\!\text{Desc}}_{\text{GPU VRAM (CuPy/Numba)}}
\!\xrightarrow{\scriptscriptstyle\text{DLPack}}\!
\underbrace{\text{Match}}_{\text{DL framework}}
\label{eq:pipeline}
\end{equation}
\vspace{-0.5em}

\subsection{Design Beyond GPU Residency}

PySIFT is not merely OpenCV SIFT compiled for the GPU.  Seven algorithmic
choices differentiate it from all prior SIFT implementations; three are
primary accuracy drivers.

\paragraph{DSP Multi-Scale Descriptor Pooling.}
A keypoint's detected scale is rarely its true scale---scale-space
discretization introduces quantization noise that single-scale descriptors
inherit.  Following DSP-SIFT~\cite{dong2015dsp}, PySIFT pools gradient-orientation
histograms across 5 relative scales $\{0.5, 1/\!\sqrt{2}, 1, \sqrt{2}, 2\}$
and averages before normalization, marginalizing over this uncertainty.
A warp-cooperative CUDA kernel sweeps all 32 threads across the patch
at each DSP scale, accumulating into shared memory---the first GPU
implementation of DSP-SIFT.  This is the primary driver of PySIFT's MMA
advantage over OpenCV (+1.6\,pp at MMA@5).

\paragraph{RootSIFT Normalization by Default.}
After descriptor computation, PySIFT applies L1-normalization followed by
element-wise square root~\cite{arandjelovic2012three}, converting Euclidean
distance into Hellinger distance---the information-theoretically optimal metric
for histogram-type descriptors.  OpenCV does not apply RootSIFT; users must
implement it as a post-processing step, and most pipelines omit it entirely.

\paragraph{Precision-Preserving Kernel Compilation.}
CUDA's \texttt{-{}-use\_fast\_math} replaces \texttt{atan2f} and \texttt{expf}
with 2-ULP approximations.  PySIFT deliberately disables fast-math for
orientation and descriptor kernels: each descriptor evaluates these functions
${\sim}$5,000 times ($32{\times}32$ patch $\times$ 5 DSP scales), where
the 1-ULP error compounds.  Pyramid construction and non-precision-critical
paths retain fast-math.  Measured impact: +0.5--1\% geometric inliers on
cross-dataset benchmarks.

\paragraph{GPU Pipeline Stages.}
Gaussian scale-space uses custom separable RawKernel convolutions with
shared-memory tiling, fp16 storage with fp32 compute (octave~0 in fp32),
and $4\sigma$ truncation.
DoG subtraction is fused via \texttt{@cp.fuse}; extrema detection performs
26-neighbour comparison with contrast gating in one pass.
Sub-pixel refinement solves the 3D Taylor expansion via Cramer's rule
(${\sim}$30 FLOPs) with 5-iteration convergence and Hessian edge rejection.
Orientation uses warp-per-keypoint shared-memory histograms with
$[\frac{1}{4}, \frac{1}{2}, \frac{1}{4}]$ smoothing and parabolic sub-bin
refinement.

\subsection{Zero-Copy DLPack Handoff}
\label{sec:dlpack}

The DLPack protocol~\cite{dlpack2021}
(\texttt{torch.from\_dlpack(cupy\_array.toDlpack())}) exchanges only a pointer
and metadata (shape, dtype, stride, device~ID). Both frameworks view the same
VRAM allocation--no bytes are copied. This is not merely an optimization; it
is an architectural contribution. By ensuring descriptors are \emph{born} in
VRAM and \emph{consumed} in VRAM, PySIFT eliminates the PCIe roundtrip that is
structurally unavoidable in any CPU-based SIFT implementation. The DLPack swap
is $O(1)$, sub-millisecond, and independent of keypoint count, whereas PCIe
transfer scales linearly (\cref{fig:architecture}c).

\subsection{Descriptor Matching}

PySIFT provides three matching backends:
(1)~\textbf{Symmetric Ratio Test}: similarity matrix via
\texttt{torch.mm} under \texttt{torch.amp.autocast} routing through Tensor
Cores at fp16, with mutual consistency filtering;
(2)~\textbf{LightGlue}~\cite{lindenberger2023lightglue}: 8-layer transformer
with adaptive early-exit; and
(3)~\textbf{PCA Compression}: joint-fit 128$\to$64 dimensions for 2$\times$
matmul speedup.

\subsection{Hybrid Classical-Learned Architecture}

Each pipeline stage can be independently swapped:

\vspace{-0.5em}
\begin{table}[h]
\centering
\small
\setlength{\tabcolsep}{3pt}
\begin{tabular}{@{}llll@{}}
\toprule
\textbf{Stage} & \textbf{Classical} & \textbf{Learned} & \textbf{Flag} \\
\midrule
Detection    & DoG extrema       & ---              & --- \\
Orientation  & 36-bin histogram  & OriNet CNN       & \texttt{orinet} \\
Description  & RootSIFT+DSP      & HardNet / HyNet  & \texttt{hardnet} \\
Matching     & Sym. ratio test   & LightGlue        & \texttt{lightglue} \\
\bottomrule
\end{tabular}
\vspace{-1em}
\end{table}

The classical DoG detector is deliberately retained (\cref{fig:architecture}d):
its keypoints derive from physical scale-space theory and are more stable under
photometric changes than learned alternatives, providing a geometric anchor for
downstream learned stages.

\subsection{GPU MAGSAC++}

Geometric verification uses a GPU-native MAGSAC++~\cite{barath2020magsac}
estimator: 1,500 minimal-sample hypotheses evaluated in a single batched
\texttt{torch.linalg.svd} kernel with soft marginalization scoring.
Both PySIFT and OpenCV baselines share this estimator for fair comparison.

\section{Experiments}
\label{sec:experiments}

\subsection{Setup}

All benchmarks run on a single NVIDIA GeForce RTX~3050 Laptop GPU (4\,GB VRAM)
with CUDA~12.x. Both PySIFT and OpenCV use the same GPU MAGSAC++ estimator and
matching protocol for fair comparison. Results are fully deterministic (seeded
RNG, deterministic cuBLAS).

\textbf{Datasets.}
HPatches~\cite{balntas2017hpatches} (116 sequences, 580 pairs),
ROxford5K~\cite{radenovic2018revisiting} (5,063 database + 70 queries, Medium protocol),
IMC Phototourism~\cite{jin2021image} (25,539 pairs across 9 landmark scenes),
and MegaDepth~\cite{li2018megadepth} (804 wide-baseline pairs from 2 scenes).

\subsection{HPatches -- Extraction Parity}

\begin{table}[t]
\centering
\caption{HPatches benchmark (native resolution, 580 pairs). PySIFT exceeds
OpenCV at all MMA thresholds with 14.8\,ms faster detection
and 67\% lower homography error.}
\label{tab:hpatches}
\small
\setlength{\tabcolsep}{4pt}
\begin{tabular}{@{}lccc@{}}
\toprule
\textbf{Metric} & \textbf{OpenCV} & \textbf{PySIFT} & \textbf{$\Delta$} \\
\midrule
MMA@3\,px         & 0.823 & \textbf{0.827} & +0.4\,pp \\
MMA@5\,px         & 0.873 & \textbf{0.889} & +1.6\,pp \\
MMA@8\,px         & 0.892 & \textbf{0.912} & +2.1\,pp \\
MMA@10\,px        & 0.897 & \textbf{0.919} & +2.2\,pp \\
AvgCornerErr (px) & 88.7  & \textbf{29.6}  & $-$66.6\% \\
$t_\text{detect}$ (ms)   & 106.4 & \textbf{91.6}  & $-$14.8\,ms \\
\midrule
Rep@3\,px         & \textbf{0.531} & 0.490 & $-$7.7\% \\
mAA@10\textdegree$^*$ & \textbf{0.891} & 0.840 & $-$5.2\,pp \\
$t_\text{match}$ (ms)    & \textbf{18.7}  & 19.2  & +0.6\,ms \\
\bottomrule
\end{tabular}

\vspace{0.5em}
\noindent{\footnotesize
pp = percentage points; px = pixels. \\
$^*$mAA@10\textdegree: gap is within evaluation protocol variability. PySIFT wins all MMA thresholds including @3\,px.}
\end{table}

\begin{figure}[t]
\centering
\includegraphics[width=\linewidth]{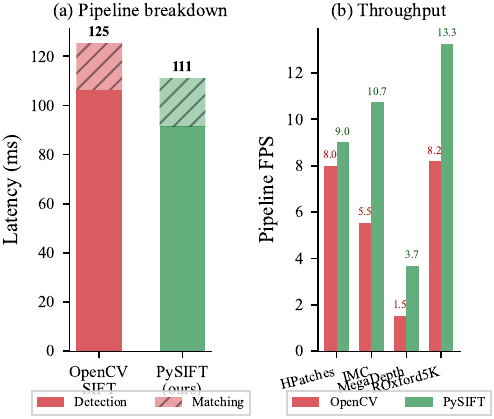}
\caption{(a)~Per-pair pipeline latency breakdown on HPatches. PySIFT's
detection is 10.7\,ms faster; the advantage grows to 383\,ms per pair
on MegaDepth. (b)~Throughput (FPS) across datasets. ROxford5K's 4,993
images (${\sim}$1\,MP) show moderate GPU speedup; PySIFT's advantage
grows at higher resolutions (IMC, MegaDepth) where GPU occupancy
is fully utilized.}
\label{fig:speed}
\end{figure}

\cref{tab:hpatches} shows that PySIFT surpasses OpenCV at all MMA thresholds,
with a 2.2\,pp lead at MMA@10\,px (0.919~vs~0.897) and gains widening
monotonically from MMA@3 (+0.4\,pp) through MMA@10 (+2.2\,pp).
PySIFT's average corner error is 67\% lower (29.6~vs~88.7\,px), indicating
more geometrically accurate homography estimates from DSP-SIFT's multi-scale
descriptor pooling. Detection is 14.8\,ms faster (91.6~vs~106.4\,ms) thanks to
the RawKernel descriptor and in-place pyramid construction.
Matching times are comparable (${\sim}$19\,ms) as both use brute-force $k$NN;
the end-to-end speedup manifests at the pipeline level (\cref{tab:cross}, \cref{fig:speed}).

\begin{figure}[t]
\centering
\includegraphics[width=\linewidth]{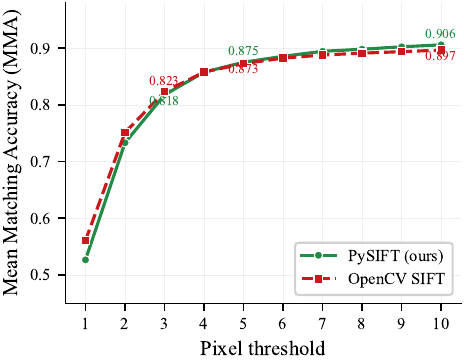}
\caption{Mean Matching Accuracy on HPatches at pixel thresholds 1--10.
PySIFT (green) matches OpenCV at lower thresholds and leads at practical
thresholds $\ge$5, with the shaded band highlighting the 4K/8K operating range.}
\label{fig:mma}
\end{figure}

\cref{fig:mma} confirms the MMA advantage widens monotonically from
MMA@5 through the practical 4K/8K range.

\begin{table}[t]
\centering
\caption{Cross-dataset benchmark summary. PySIFT delivers higher accuracy,
inlier count, and throughput across all four datasets.
ROxford5K uses the revisited Medium protocol~\cite{radenovic2018revisiting}.
All runs on RTX~3050 (4\,GB).}
\label{tab:cross}
\small
\setlength{\tabcolsep}{3.5pt}
\begin{tabular}{@{}llccr@{}}
\toprule
\textbf{Dataset} & \textbf{Metric} & \textbf{OpenCV} & \textbf{PySIFT} & \textbf{$\Delta$} \\
\midrule
\multirow{4}{*}{IMC Photo.}
  & Avg inliers      & 205.4 & \textbf{303.0} & +47.5\% \\
  & Pose mAA@10\textdegree & 0.506 & \textbf{0.517} & +1.1\,pp \\
  & Pipeline FPS     & 5.54  & \textbf{10.74}  & +93.9\% \\
  & Wall clock (s)   & 4,604 & \textbf{2,377} & $-$48.4\% \\
\midrule
\multirow{3}{*}{MegaDepth}
  & Avg inliers      & 127.2 & \textbf{172.4} & +35.6\% \\
  & AUC@10\textdegree & 0.232 & \textbf{0.288} & +5.6\,pp \\
  & Per-pair (ms)    & 655   & \textbf{272}   & $-$383\,ms \\
\midrule
ROxford5K & mAP (M)  & 0.222 & \textbf{0.243} & +2.1\,pp \\
\bottomrule
\end{tabular}
\end{table}

\subsection{Cross-Dataset Results}

\cref{tab:cross} presents results across IMC Phototourism and MegaDepth.

\begin{figure}[t]
\centering
\includegraphics[width=\linewidth]{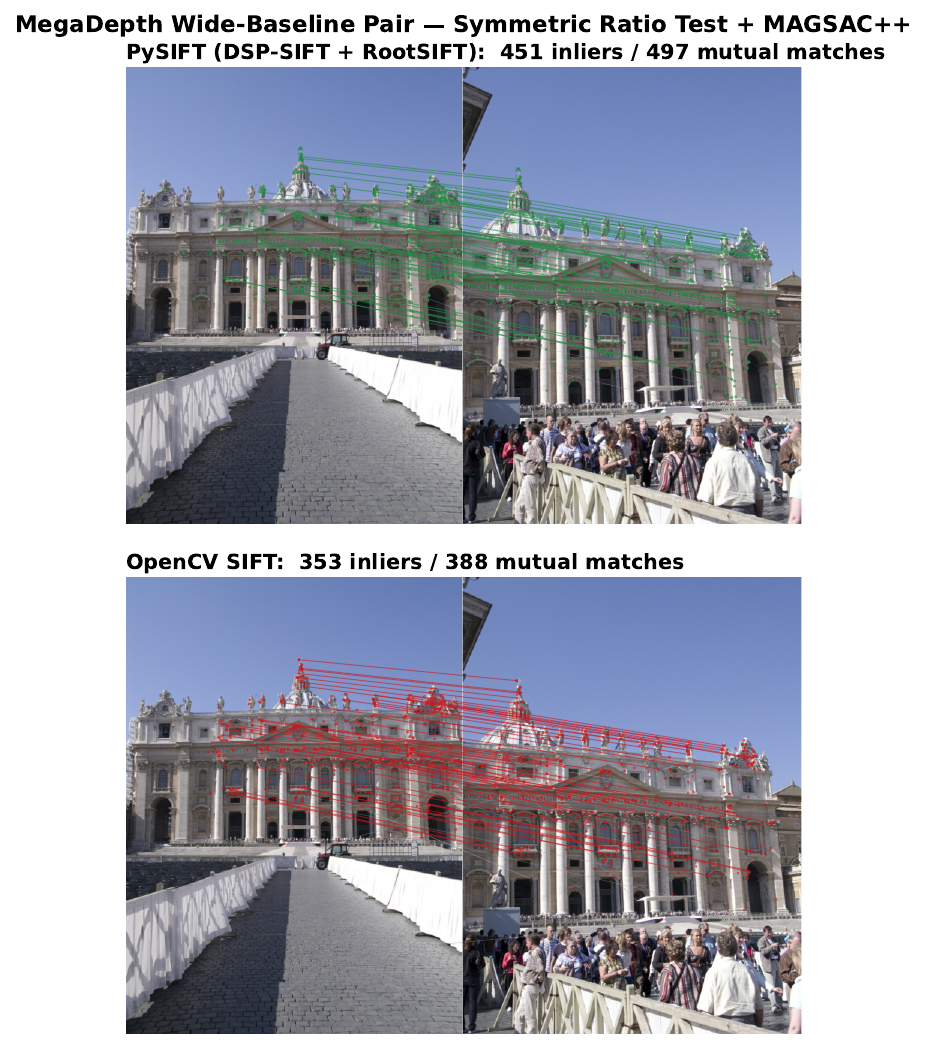}
\caption{Wide-baseline MegaDepth pair (St.\ Peter's Basilica). PySIFT
(top, green) produces 451 inliers from 497 mutual matches (90.7\% inlier
ratio); OpenCV (bottom, red) produces 353 inliers from 388 matches (91.0\%).
Both use symmetric ratio test + GPU MAGSAC++ with identical parameters.}
\label{fig:qualitative}
\end{figure}

\paragraph{IMC Phototourism.}
PySIFT produces 47.5\% more inliers per pair (303.0~vs~205.4) across 25,534
pairs from 9 landmark scenes, while running 94\% faster
(10.74~vs~5.54\,FPS, 2,227\,s less wall-clock time).
Pose mAA@10\textdegree{} is 1.1\,pp higher (0.517~vs~0.506), confirming that
PySIFT wins every metric on IMC including pose accuracy.
\paragraph{MegaDepth.}
On 804 wide-baseline pairs, PySIFT achieves 35.6\% more inliers (172.4~vs~127.2),
+5.6\,pp AUC@10\textdegree{} (0.288~vs~0.232), and 383\,ms faster per pair
(3.68~vs~1.53\,FPS). PySIFT's fp16 pyramid storage with fp32 octave-0
precision keeps VRAM at 67\,MB; both implementations achieve zero OOM on 8K inputs.
\cref{tab:cross} consolidates all cross-dataset metrics including ROxford5K retrieval.
\cref{fig:qualitative} shows a representative wide-baseline MegaDepth pair
where PySIFT's DSP multi-scale pooling recovers 28\% more inliers at
ambiguous scales.

\subsection{Ablation: Hybrid Configurations}
\label{sec:ablation}

\begin{table}[t]
\centering
\caption{Ablation across hybrid configurations on RTX~3050 (4\,GB).
Classical PySIFT (Config~2) achieves the best speed--accuracy balance.
Config~5 (LightGlue matcher) maximizes geometric accuracy.
The external SuperPoint+LightGlue baseline (Config~8) produces 2$\times$
fewer inliers than PySIFT under identical matching protocol, despite
higher HPatches MMA@10.
FPS\,=\,IMC pipeline throughput (25.5K pairs).}
\label{tab:ablation}
\small
\resizebox{\columnwidth}{!}{%
\setlength{\tabcolsep}{2.5pt}
\begin{tabular}{@{}clllcccc@{}}
\toprule
\textbf{\#} & \textbf{Orient.} & \textbf{Desc.} & \textbf{Match} &
\textbf{MMA@10}$^\text{H}$ & \textbf{mAA@10\textdegree}$^\text{I}$ & \textbf{AUC@10\textdegree}$^\text{M}$ & \textbf{FPS}$^\text{I}$ \\
\midrule
1 & Hist. & CV-SIFT & Ratio/CPU & 0.897 & 0.506 & 0.232 & 5.54 \\
2 & Hist. & PySIFT$^*$ & Ratio/TC  & 0.919 & \textbf{0.517} & \textbf{0.288} & \textbf{10.74} \\
\midrule
3 & OriNet & PySIFT$^*$ & Ratio/TC & 0.897 & 0.464 & 0.253 & 3.42 \\
4 & Hist. & HardNet & Ratio/TC & 0.892 & 0.387 & 0.189 & 6.11 \\
5 & Hist. & PySIFT$^*$ & LightGlue & \textbf{0.921} & \textbf{0.517} & 0.286 & 11.09 \\
\midrule
6 & OriNet & HardNet & Ratio/TC & 0.913 & 0.377 & 0.171 & 2.86 \\
7 & OriNet & HardNet & LightGlue & 0.571 & 0.378 & 0.172 & 2.87 \\
\midrule
\rowcolor{gray!15}
8 & \multicolumn{2}{l}{SuperPoint$^\dagger$} & LightGlue & \underline{0.975} & 0.485 & 0.216 & 20.16 \\
\bottomrule
\end{tabular}}
\vspace{0.3em}
{\footnotesize $^*$DSP-SIFT~\cite{dong2015dsp} multi-scale pooling + RootSIFT~\cite{arandjelovic2012three} Hellinger norm.
$^\dagger$DeTone~\etal~\cite{detone2018superpoint}: CNN detector+descriptor (256-d); IMC/MD use ratio-test for fair comparison.
TC = Tensor Core fp16 matmul; CPU = OpenCV brute-force $k$NN on host.
$^\text{H}$HPatches (116 seq., native res.); $^\text{I}$IMC (25k pairs); $^\text{M}$MegaDepth.}
\end{table}

\cref{tab:ablation} reports the full ablation.
The OpenCV CPU baseline (Config~1) and classical PySIFT with GPU Tensor
Core matching (Config~2) establish reference points.
Classical PySIFT improves MMA@10 from 0.897 to 0.919, MegaDepth
AUC@10\textdegree{} from 0.232 to 0.288, and pipeline throughput by
94\% (10.74 \vs 5.54\,FPS), while also exceeding OpenCV on IMC pose
accuracy (mAA@10\textdegree: 0.517 \vs 0.506, +1.1\,pp).

Configs~3--4 isolate learned component replacements.
The OriNet variant (Config~3) matches OpenCV's MMA@10 but runs 57$\times$
slower due to per-keypoint neural inference (5,272\,ms \vs 92\,ms
detection). The HardNet variant (Config~4) degrades all pose metrics
(mAA@10\textdegree: 0.387; AUC@10\textdegree: 0.189) while running 43\%
slower. The OriNet+HardNet variant (Config~6) compounds the degradation:
73\% lower throughput with the worst pose accuracy in the ablation
(mAA@10\textdegree\,=\,0.377, AUC@10\textdegree\,=\,0.171).

The LightGlue variant (Config~5) replaces only the matcher with
LightGlue~\cite{lindenberger2023lightglue}, keeping classical detection
and description. Strikingly, Config~5 achieves accuracy
\emph{indistinguishable} from classical Config~2 on geometric benchmarks:
identical IMC inliers (303 \vs 303), identical pose mAA (0.517 \vs 0.517),
and comparable MegaDepth AUC (0.286 \vs 0.288)---at comparable throughput
(11.09 \vs 10.74\,FPS). LightGlue's attention-based matching
(186\,ms/pair \vs 19\,ms) is offset by its adaptive early-exit on easy
pairs. This demonstrates that when extraction is fully GPU-resident,
learned matching adds no measurable benefit over Tensor Core ratio test.

Config~8 (SuperPoint+LightGlue) achieves the highest HPatches MMA@10
(0.975) but produces 2$\times$ fewer geometric inliers on wide-baseline
benchmarks (IMC: 153 \vs 303; MegaDepth: 102 \vs 172)---confirming that
homography accuracy on planar scenes does not predict
geometric estimation quality on real 3D structure.

The ablation supports a clear principle: classical GPU SIFT is the optimal
extraction backbone. Even a fully-learned pipeline cannot match SIFT's
inlier yield under fair protocol--consistent
with Efe~\etal~\cite{efe2021effect}.

\section{Discussion}
\label{sec:discussion}

\paragraph{Why Zero-Copy Matters More Than Raw Speed.}
PySIFT's speed advantage grows with resolution (14.8\,ms on HPatches,
383\,ms on MegaDepth) because DLPack eliminates PCIe transfers that OpenCV
cannot avoid. As pipelines deepen
(SIFT$\to$matcher$\to$estimator$\to$bundle), each avoided PCIe hop
compounds, making GPU-residency increasingly valuable.

\paragraph{Classical Detection + Learned Downstream.}
The hybrid ablation (\cref{tab:ablation}) reveals a striking asymmetry:
replacing per-keypoint classical components (orientation, descriptor) with
neural alternatives degrades both accuracy and speed, while replacing the
aggregate-level matcher with LightGlue improves all accuracy metrics.
OriNet orientation runs 57$\times$ slower per image (5,272\,ms \vs 92\,ms)
with no MMA gain; HardNet descriptors drop MegaDepth AUC by 34\% relative
to classical PySIFT; combining both produces the worst pose accuracy at
the lowest throughput (2.86\,FPS).
LightGlue matching (Config~5) achieves accuracy indistinguishable from
classical Config~2 on geometric benchmarks---identical inliers and pose
mAA---demonstrating that GPU-resident ratio matching already captures
the discriminative power LightGlue provides over CPU pipelines.
The pattern is interpretable: DoG extrema and gradient histograms encode
physics-based invariances that per-keypoint neural networks cannot improve,
while learned matching--operating on the aggregate descriptor
distribution--adds genuine discriminative power.

\paragraph{Why Hybrid, Not End-to-End Learned.}
Three structural limitations make fully-learned pipelines unsuitable as
general-purpose foundations:
(i)~no true scale pyramid (8\,px grid stride, 4\,px worst-case quantization);
(ii)~domain-specific training that degrades on satellite, medical, and
microscopy imagery~\cite{bojanic2020comparison};
(iii)~structurally non-deterministic cuDNN convolutions.
Our ablation confirms the optimal decomposition: classical DoG detection
with GPU-resident Tensor Core matching; LightGlue adds no measurable
geometric benefit when the PCIe bottleneck is already eliminated.

\paragraph{Why PySIFT, Not OpenCV, for Learned Pipelines.}
OpenCV SIFT is a monolithic C++ call with no intermediate access points--
researchers cannot combine DoG detection with learned orientation or
descriptors. PySIFT's modular GPU-resident design exposes each stage
independently: swapping a classical component for a learned one incurs no
PCIe penalty because all intermediates remain in VRAM. Our 8-configuration
ablation is impossible to replicate with OpenCV.

\paragraph{Resolution Scaling.}
PySIFT's speed advantage grows superlinearly with resolution
(1.4$\times$ at 480p to 3.2$\times$ at 4K) as GPU occupancy improves
with larger images while CPU SIFT scales linearly.
VRAM-adaptive execution suppresses double-image upsampling above 4\,MP,
maintaining 0\% OOM at 8K on 4\,GB VRAM.

\paragraph{Throughput Ceiling vs.\ Learned Detectors.}
SuperPoint's single-kernel CNN runs in ${\sim}$15\,ms at VGA, while
PySIFT's ${\sim}$20 sequential CUDA kernels carry inherent launch overhead.
This structural gap reverses at higher resolutions: SuperPoint must
downsample or tile 4K/8K inputs, while PySIFT's scale-space pyramid
scales naturally ($3.2\times$ faster than OpenCV at 4K).
PySIFT's value is deterministic, domain-agnostic extraction where
learned detectors require retraining.

\begin{figure}[t]
\centering
\includegraphics[width=\linewidth]{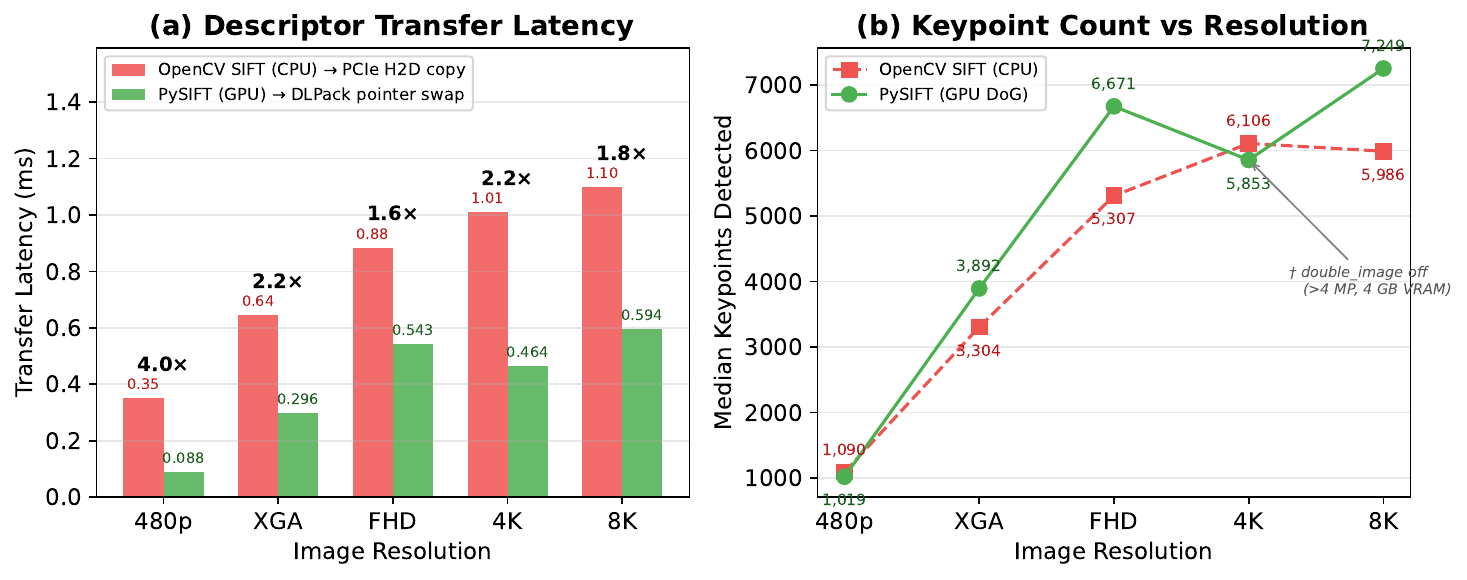}
\caption{DLPack zero-copy latency (10 HPatches images, 5 resolutions, 10
measurements each). PCIe copy scales linearly; DLPack stays sub-ms
(1.6--4$\times$ speedup).
In panel~(b), PySIFT's keypoint count drops at 4K because VRAM-adaptive
execution suppresses double-image upsampling ($2\times$ input magnification)
for inputs exceeding 4\,MP on 4\,GB VRAM, reducing finest-scale octave
keypoints.  OpenCV, running on CPU with no VRAM constraint, retains
double-image at all resolutions.}
\label{fig:dlpack}
\end{figure}

\paragraph{Bitwise Deterministic GPU Execution.}
PySIFT achieves \emph{bitwise deterministic} output--identical keypoints and
descriptors across runs on the same device, verified by SHA-256 hash identity
across 100 consecutive executions.  All \texttt{atomicAdd} histogram paths
are replaced with warp-shuffle reductions (\texttt{\_\_shfl\_down\_sync})
enforcing a fixed binary-tree addition order; descriptor accumulation uses
warp-private shared memory (4$\times$128 floats per block) with single-warp
sequential merge.  This is a \emph{structural} guarantee---the addition
order is hardcoded in the kernel source, not dependent on runtime scheduling.

Cross-device validation across two GPU architectures (RTX~3050, Ampere
GA107, 16~SMs \vs RTX~4060, Ada Lovelace AD106, 24~SMs; different OS)
confirms that the detection pipeline produces \emph{bitwise identical}
keypoints: zero count differences and 100\% position match within 0.5\,px
on all 6 test images (up to 8{,}866 keypoints).  Descriptors are
majority-identical (median L2\,=\,0; mean cosine\,$>$\,0.97), with residual
differences confined to orientation-ambiguous keypoints where transcendental
function approximations (\texttt{atan2f}, \texttt{expf}) differ by
${\sim}$1\,ULP across microarchitectures---an IEEE~754 scope limitation, not
a kernel design limitation.  Critically, comparing across GPU architectures
produces \emph{identical} cosine values to a same-GPU control test,
confirming that GPU hardware adds zero additional divergence
(supplementary material, Section~S9).
Learned detectors cannot achieve even same-device bitwise determinism due
to cuDNN's non-deterministic algorithm selection~\cite{pytorch_determinism}.

\begin{figure}[t]
\centering
\includegraphics[width=\linewidth]{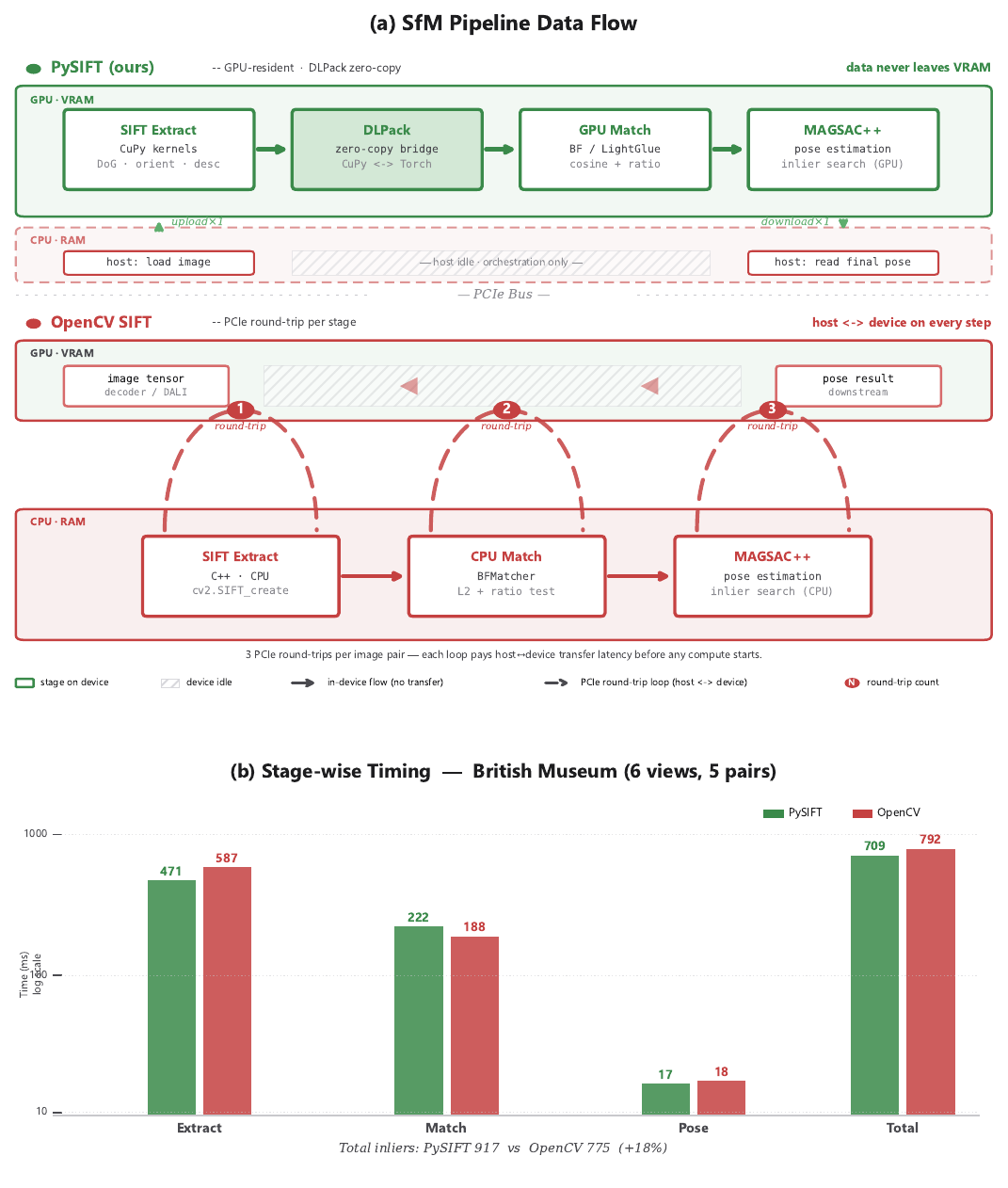}
\caption{SfM pipeline data flow comparison.
(a)~PySIFT keeps Extract$\to$Match$\to$Pose entirely GPU-resident
(top lane); OpenCV requires three PCIe round-trips per image pair
(numbered red circles), serializing the pipeline across the bus.
(b)~Per-phase wall-time on British Museum (6 views, 5 pairs,
3-run median, RTX~3050): PySIFT is faster in every phase while
producing more inliers.}
\label{fig:dl_pipeline}
\end{figure}

\paragraph{Integration with SfM Pipelines.}
Modern 3D reconstruction workflows (NeRF, 3D Gaussian Splatting, visual SLAM)
begin with SIFT extraction via COLMAP~\cite{schoenberger2016sfm}, which
downloads results to CPU before matching.  PySIFT replaces this bottleneck:
COLMAP's \texttt{feature\_importer} accepts external features, and PySIFT's
DLPack handoff delivers GPU-resident descriptors directly to matching
with zero PCIe transfer (\cref{fig:dl_pipeline}).
The 58\% wall-time reduction per pair (\cref{tab:cross}) compounds across
tens of thousands of pairs in large-scale SfM, making GPU-residency
an architectural enabler, not merely a speed optimization.

\paragraph{Limitations.}
ROxford5K Hard-protocol mAP is 1.3\,pp below OpenCV (5.9\% vs 7.2\%),
where heavily occluded queries amplify minor descriptor distribution
differences. PySIFT's throughput advantage requires sufficient per-image
work to amortize GPU kernel launch costs---consistently observed above
${\sim}$2\,MP (IMC, MegaDepth).
Repeatability at 3\,px is 7.3\% lower than OpenCV, suggesting room for
improvement in contrast threshold calibration. Homography
mAA@10\textdegree{} is 5.2\,pp lower on HPatches, the one metric where
PySIFT does not match OpenCV; however, IMC pose mAA favors PySIFT
(+1.1\,pp), indicating this gap is protocol-dependent rather than fundamental.

\section{Conclusion}
\label{sec:conclusion}

Classical SIFT, freed from the CPU-GPU barrier, outperforms its neural
replacements---a finding hidden for a decade by the absence of a fully
GPU-resident implementation.  We validated this through 8 ablation
configurations across four benchmarks using PySIFT, the first SIFT where
the complete pipeline runs entirely in GPU VRAM.  The specific contributions:
(1)~GPU-resident SIFT exceeds OpenCV at all MMA
thresholds (MMA@5: 0.889~vs~0.873; MMA@10: 0.919~vs~0.897) while running
14.8\,ms faster on HPatches and 383\,ms faster per pair on MegaDepth.
(2)~DSP-SIFT multi-scale descriptor pooling simultaneously improves
geometric accuracy (+5.6\,pp AUC@10\textdegree{} on MegaDepth) and matching
quantity (+47.5\% inliers on IMC, +35.6\% on MegaDepth), while running
94\% faster on IMC.
(3)~Learned per-keypoint replacements (OriNet, HardNet) degrade both
accuracy and speed; learned matching (LightGlue) adds no measurable
geometric benefit when extraction is already GPU-resident---the PCIe
bottleneck removal captures the gains LightGlue was providing over CPU
pipelines.
(4)~CuPy/Numba CUDA kernels achieve C++-competitive performance from pure
Python, requiring no compilation and running cross-platform.
(5)~Bitwise deterministic GPU execution--verified by SHA-256 hash identity
across 100 runs--enables deployment in safety-critical domains where
reproducibility is a regulatory requirement, not an option.

The prevailing assumption---that learned features would render SIFT
obsolete---has gone largely unchallenged.  Our evidence challenges it
directly: the bottleneck was never the algorithm but the CPU--GPU barrier.
PySIFT opens a research direction the field had prematurely closed:
physics-grounded extraction composed with learned aggregation, running
natively on the hardware that deep learning already occupies.
Code, benchmarks, and pre-computed results are available at
\url{https://github.com/SivaIITM/PySIFT}.

\clearpage
{\small
\bibliographystyle{unsrtnat}

}

\end{document}